\documentclass[11pt]{article}

\usepackage[utf8]{inputenc}
\usepackage[T1]{fontenc}
\usepackage{amsmath,amssymb}
\usepackage{graphicx}
\usepackage{geometry}
\usepackage{authblk}
\usepackage{hyperref}
\usepackage{cases}
\usepackage{xcolor}
\usepackage[normalem]{ulem}

\title{Prediction of Nonlinear Oscillations in a Jumping Quarter-Car Model Using Reservoir Computing}

\author[a]{Masahisa Watanabe\thanks{Corresponding author: masahisa-watanabe@go.tuat.ac.jp}}
\author[b]{Shiva Dixit}
\author[b]{Nirmal Punetha}
\author[c]{Swati Chauhan}
\author[d]{Manish Dev Shirimali}

\affil[a]{Division of Environmental and Agricultural Engineering, Institute of Agriculture, Tokyo University of Agriculture and Technology, Tokyo, Japan}
\affil[b]{Amity Institute of Integrative Sciences and Health, Amity University Haryana - Gurugram, Manesar, 122 413 Haryana, India}
\affil[c]{Graduate School of Engineering, Nagoya Institute of Technology, Nagoya 466 8555, Japan}
\affil[d]{Department of Physics, Central University of Rajasthan, Ajmer, 305 817 Rajasthan, India}

\date{}

\begin{document}

\maketitle

\noindent ORCID ID (Masahisa Watanabe): \url{https://orcid.org/0000-0002-1623-0033}

\begin{abstract}
\noindent Reliable prediction of vehicle dynamics is essential for smart driving applications such as autonomous control and advanced driver-assistance systems. Off-road vehicles used in agricultural and construction settings are particularly prone to nonlinear behavior, including bifurcations and chaotic motion arising from intermittent loss of tire--road contact. Predicting such dynamics is challenging because it requires resolving both smooth nonlinearities and the discontinuous switching associated with contact loss. In this work, we investigate the feasibility of reservoir computing (RC) -- specifically an echo state network (ESN) -- for data-driven prediction of a jumping quarter-car model. The reservoir is trained on time-series data from a small number of points and evaluated on its ability to reconstruct bifurcation diagrams, phase-space attractors, and time trajectories across periodic and chaotic regimes. The trained reservoir qualitatively reproduces the period-doubling route to chaos, captures the geometric structure of periodic and chaotic attractors. These results demonstrate that reservoir computing is a feasible data-driven predictor of nonlinear dynamics in a practical, non-smooth vehicle system.

\vspace{0.5em}
\noindent \textbf{Keywords:} Quarter-car model; Reservoir computing; Chaos prediction; Echo state network; Bifurcation Analysis
\end{abstract}

\noindent \textbf{Acknowledgment}

\noindent This work was supported by Japan Society for the Promotion of Science Grants-in-Aid (26K14043).

\section*{Nomenclature and Abbreviations}

\begin{center}
\begin{tabular}{ll}
\hline
Symbols & Meaning of symbols \\
\hline
$z_s$ & Vertical motion of sprung mass [m] \\
$z_u$ & Vertical motion of unsprung mass [m] \\
$Z_s$ & Suspension stroke [m] \\
$Z_u$ & Tire deflection [m] \\
$f_s$ & Suspension force [N] \\
$f_t$ & Tire force [N] \\
$m_s$ & Sprung mass [kg] \\
$m_u$ & Unsprung mass [kg] \\
$k_s$ & Suspension stiffness [N m$^{-1}$] \\
$k_t$ & Tire stiffness [N m$^{-1}$] \\
$k_{to}$ & Original (contact) value of tire stiffness [N m$^{-1}$] \\
$c_s$ & Suspension damping coefficient [N s m$^{-1}$] \\
$g$ & Gravitational acceleration [m s$^{-2}$] \\
$d_0$ & Road excitation amplitude [m] \\
$t$ & Time [s] \\
$t_s$ & Scaled (dimensionless) time [rad] \\
$\omega$ & Road angular excitation frequency [rad s$^{-1}$] \\
$\omega_s$ & Natural angular frequency of the sprung mass [rad s$^{-1}$] \\
$\alpha$ & Mass ratio, $m_s/m_u$ [-] \\
$\beta$ & Stiffness ratio, $k_t/k_s$ [-] \\
$\beta_j$ & Switched (jumping) value of the stiffness ratio [-] \\
$\zeta$ & Damping ratio of the sprung mass [-] \\
$\gamma$ & Dimensionless gravity [-] \\
$\Omega$ & Frequency ratio, $\omega/\omega_s$ [-] \\
$N_{rc}$ & Number of reservoir nodes [-] \\
$\alpha_{rc}$ & Reservoir leak rate [-] \\
$\rho_{rc}$ & Reservoir spectral radius [-] \\
$k_{rc}$ & Average reservoir connectivity degree [-] \\
$W_{in}$ & Input scaling coefficient [-] \\
$\beta_{rc}$ & Ridge regression regularization coefficient [-] \\
\hline
\end{tabular}
\end{center}

\section{Introduction}

Prediction of vehicle dynamics is significantly important to achieve smart driving, such as autonomous driving or advanced driver-assistance systems. Thus far, several studies have been conducted on the prediction of vehicle behavior \cite{ref1,ref2}. These studies mostly assume a linear vehicle model. In reality, vehicle dynamics can demonstrate rich nonlinearities, such as nonlinear axle suspensions \cite{ref3}, steering systems \cite{ref4}, and traction \cite{ref5}. These nonlinearities induce significant dynamic instabilities, such as bifurcation and chaos, which can deteriorate the stability and controllability of the vehicle system. Especially in off-road vehicle operation, such as agricultural and construction applications, dynamic instabilities can substantially reduce operational efficiency and, in severe cases, lead to hazardous accidents \cite{ref6,ref7}. Therefore, prediction of nonlinear vehicle behavior, especially chaos prediction, is necessary for smart driving.

Considerable research has been conducted on chaos prediction \cite{ref8,ref9,ref10}. Classical approaches to this problem generally rely on local linear approximations of the underlying dynamics, which can be sensitive to model mismatch and require accurate knowledge of the governing equations. Owing to advances in computational resources and learning theory, machine learning techniques have increasingly been used for chaos prediction \cite{ref11} as a data-driven alternative that does not require an explicit dynamical model. In particular, reservoir computing has been actively investigated and applied to the prediction of nonlinear dynamical systems \cite{ref12,ref13,ref14,ref15}. However, applications of reservoir computing to chaos prediction have mainly focused on theoretical dynamical systems, and few studies have examined its use for practical industrial systems that combine smooth nonlinearities with discontinuous, contact-induced switching. The objective of this short report is to demonstrate, as a feasibility study, whether reservoir computing can predict the bifurcation structure, attractor topology, and time-domain behavior of a jumping quarter-car model -- a practical vehicle system exhibiting exactly this type of non-smooth nonlinearity.

\subsection{Vehicle model}

\begin{figure}[htbp]
\centering
\includegraphics[width=0.4\textwidth]{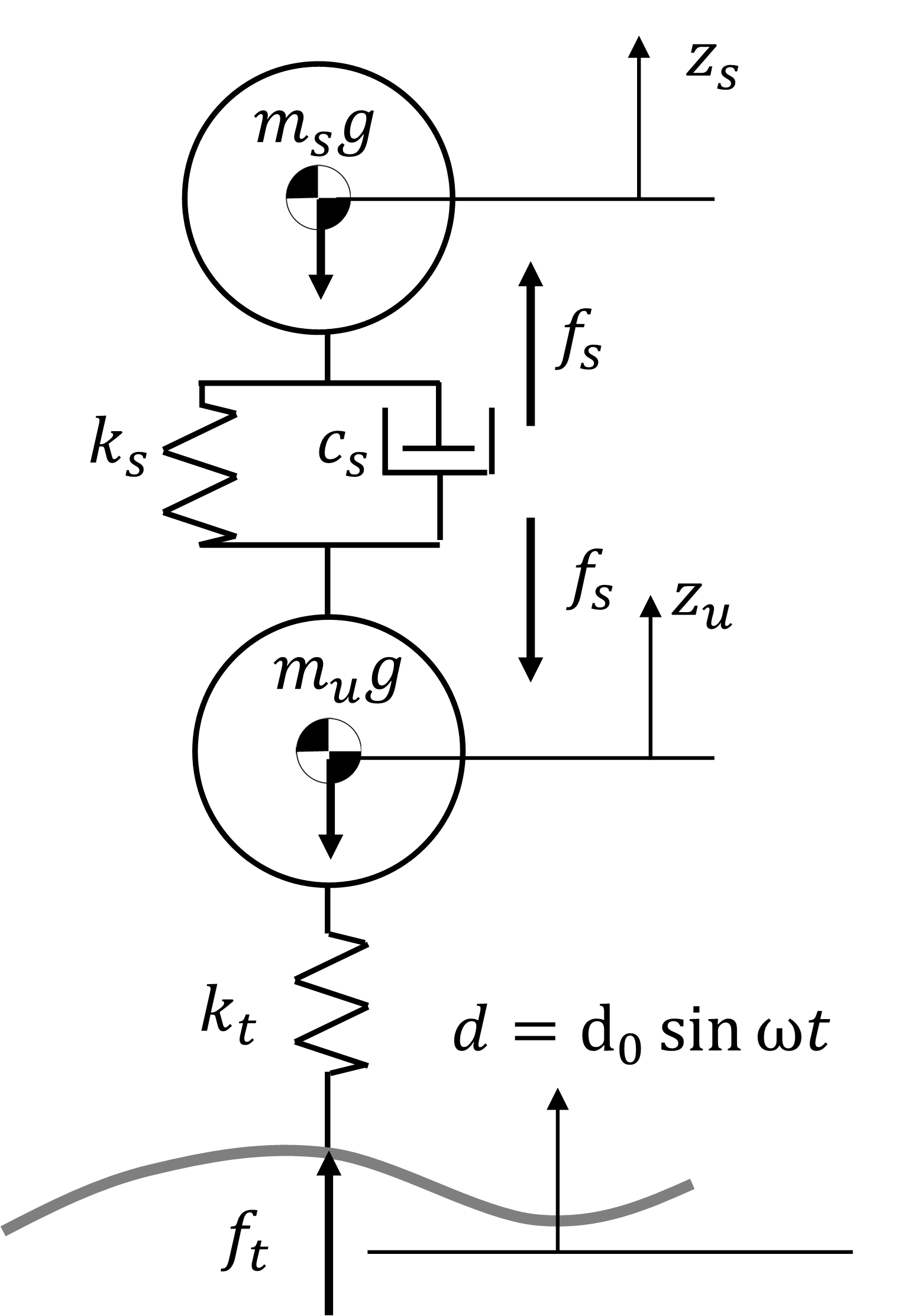}
\caption{Schematic representation of the quarter-car model incorporating a jumping (contact-loss) nonlinearity. The sprung mass $m_s$ and unsprung mass $m_u$ are coupled through a linear suspension spring--damper pair ($k_s$, $c_s$), while the unsprung mass interacts with the road profile $d(t) = d_0\sin\omega t$ through a tire modeled as a unilateral spring of stiffness $k_t$. The tire stiffness is switched to zero whenever the wheel loses contact with the road surface, giving rise to the piecewise-smooth governing dynamics.}
\label{fig:fig1}
\end{figure}

In this section, the nonlinear vehicle model used in this study is explained. A quarter-car model with jumping nonlinearity is used because jumping is quite common in off-road vehicle dynamics \cite{ref17,ref18,ref19}. Figure~\ref{fig:fig1} shows a schematic of the model when the vehicle maintains and loses contact with the supporting road.

The equations of motion are as follows:

\begin{equation}
m_s \ddot{z}_s = f_s - m_s g,
\end{equation}

\begin{equation}
m_u \ddot{z}_u = f_t - f_s - m_u g,
\end{equation}

where the dot denotes differentiation with respect to time $t$, and $f_s$ and $f_t$ denote the passive suspension force and the vertical tire force, respectively. These forces are expressed as follows:

\begin{equation}
f_s = -k_s\left(z_s - z_u\right) - c_s\left(\dot{z}_s - \dot{z}_u\right),
\end{equation}

\begin{equation}
f_t = -k_t(z_u - d),
\end{equation}

where $k_s$, $k_t$, and $c_s$ denote the suspension stiffness, tire stiffness, and suspension damping coefficient, respectively. If the relative displacement $z_u - d$ becomes positive, the tire loses contact with the ground. In this case, the tire no longer acts as a spring, and the tire force is zero. To describe this nonlinearity, similar to bouncing-ball dynamics, the tire stiffness is switched depending on the sign of $z_u - d$, as follows:

\begin{equation}
k_t =
\begin{cases}
k_{to}, & z_u - d < 0 \\
0, & z_u - d \geq 0
\end{cases}
\end{equation}

A sinusoidal road profile is used:

\begin{equation}
d(t) = d_0 \sin{\omega t}.
\end{equation}

The following dimensionless variables and parameters are introduced to derive the nondimensional form of the model:

\begin{equation}
Z_s = z_s - z_u,
\end{equation}

\begin{equation}
Z_u = z_u - d,
\end{equation}

\begin{equation}
t_s = \omega_s t,
\end{equation}

\begin{equation}
\omega_s = \sqrt{\frac{k_s}{m_s}},
\end{equation}

\begin{equation}
\alpha = \frac{m_s}{m_u},
\end{equation}

\begin{equation}
\beta = \frac{k_t}{k_s},
\end{equation}

\begin{equation}
\zeta = \frac{c_s}{2\sqrt{m_s k_s}},
\end{equation}

\begin{equation}
\gamma = \frac{g}{\omega_s^2} = \frac{m_s g}{k_s},
\end{equation}

\begin{equation}
\Omega = \frac{\omega}{\omega_s} = \omega\sqrt{\frac{m_s}{k_s}}.
\end{equation}

Using the variables and parameters above, the dimensionless equations of motion are derived as

\begin{equation}
\ddot{Z}_s = -(1+\alpha)Z_s - 2\zeta(1+\alpha)\dot{Z}_s + \alpha\beta Z_u,
\label{eq:Zs_eom}
\end{equation}

\begin{equation}
\ddot{Z}_u = \alpha Z_s + 2\zeta\alpha\dot{Z}_s - \alpha\beta Z_u - \gamma + d_0\Omega^2\sin{\Omega t_s},
\label{eq:Zu_eom}
\end{equation}

where the dot denotes differentiation with respect to $t_s$. To describe the jumping nonlinearity of the vehicle in Eqs.~(\ref{eq:Zs_eom})--(\ref{eq:Zu_eom}), $\beta$ is switched depending on $Z_u$:

\begin{equation}
\beta =
\begin{cases}
\beta_j, & Z_u < 0 \\
0, & Z_u \geq 0
\end{cases}
\label{eq:beta_switch}
\end{equation}

\noindent where $\beta_j = k_{to}/k_s$ is the stiffness ratio evaluated at the original (in-contact) tire stiffness.

\subsection{Reservoir computing}

Training a fully recurrent neural network (RNN) is not straightforward in practice. The standard approach, backpropagation through time, tends to converge slowly and is prone to vanishing or exploding gradients when sequences are long \cite{ref12}. Reservoir computing avoids this problem by not training the recurrent part of the network at all. Instead, a large recurrent network -- the ``reservoir'' -- is generated once, randomly, and then left fixed; only a simple linear readout from the reservoir states to the desired output is trained, reducing the learning problem to an ordinary linear regression \cite{ref12,ref13}. This idea was introduced independently under different names -- echo state networks (ESN) by Jaeger \cite{ref12} and liquid state machines (LSM) by Maass et al. \cite{ref13} -- and was shown early on to be remarkably effective: Jaeger and Haas demonstrated that an ESN could outperform previous methods on a standard chaotic time-series benchmark by several orders of magnitude \cite{ref16}, which largely drew attention to the approach.

For a reservoir to be useful, it must satisfy the echo state property: the influence of the initial reservoir state should die out over time, so that the current reservoir state depends only on recent input history rather than on initialization \cite{ref12}. In practice, this is achieved by scaling the reservoir's recurrent weight matrix so that its spectral radius $\rho_{rc}$ remains close to or below one; if $\rho_{rc}$ is set too low the reservoir forgets useful history too quickly, while if it is set too high the reservoir state fails to settle and training becomes unstable.

In this study, an ESN is used as the reservoir. At every time step, the current state of the quarter-car model -- $Z_s$, $Z_u$, and their time derivatives -- together with the road excitation amplitude $d_0$, is fed into the reservoir through a sparse, randomly weighted input connection with scaling $W_{in}$. The reservoir itself is a sparsely connected network (average connectivity degree $k_{rc}$) of $N_{rc}$ nonlinear (tanh) units with leak rate $\alpha_{rc}$, following the leaky-integrator formulation commonly used in ESN implementations. The readout weights connecting reservoir states to the predicted output are trained offline using ridge regression with regularization coefficient $\beta_{rc}$, chosen to avoid overfitting to the relatively short training trajectories used here.

\begin{figure}[htbp]
\centering
\includegraphics[width=0.75\textwidth]{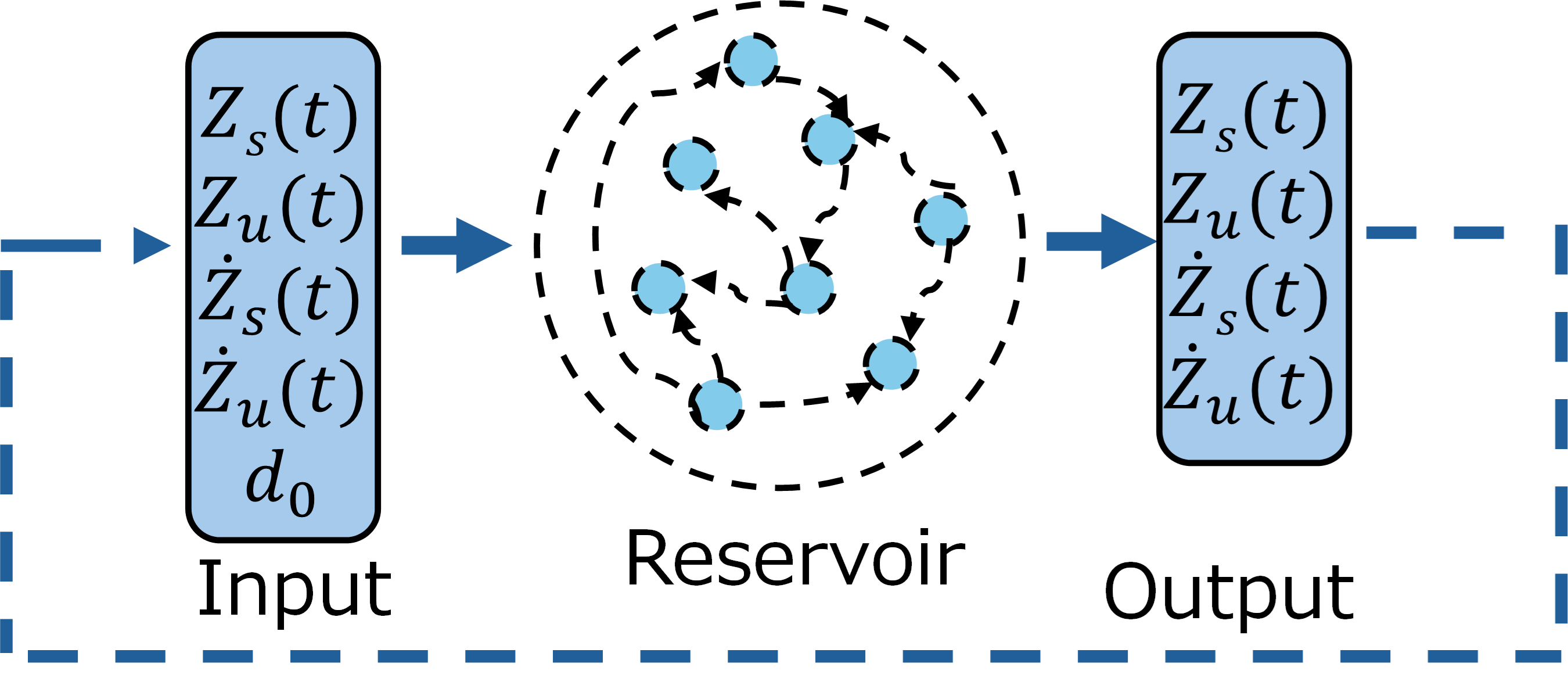}
\caption{Schematic diagram of the echo state network (ESN) architecture employed as the reservoir-computing predictor. The input layer supplies the instantaneous state vector of the quarter-car model together with the road excitation amplitude $d_0$ to a fixed, sparsely and randomly connected reservoir of nonlinear units. Only the linear readout mapping from the reservoir state to the predicted output is trained, via ridge regression, while the internal reservoir weights remain untrained throughout.}
\label{fig:fig2}
\end{figure}

Formally, the reservoir state $\mathbf{r}(t) \in \mathbb{R}^{N_{rc}}$ is updated at each time step according to
\begin{equation}
\mathbf{r}(t+\Delta t) = (1-\alpha_{rc})\,\mathbf{r}(t) + \alpha_{rc}\tanh\left(W_{res}\,\mathbf{r}(t) + W_{in}\,\mathbf{u}(t)\right),
\label{eq:reservoir_update}
\end{equation}
where $\mathbf{u}(t) = \left[Z_s, Z_u, \dot{Z}_s, \dot{Z}_u, d_0\right]^\top$ is the input vector, $W_{res}$ is the fixed, sparse, randomly generated reservoir weight matrix with spectral radius $\rho_{rc}$ and connectivity degree $k_{rc}$, and $W_{in}$ is the fixed, randomly generated input weight matrix whose elements are randomly generated from a uniform distribution. The predicted output is obtained from a linear readout,
\begin{equation}
\hat{\mathbf{y}}(t) = W_{out}\,\mathbf{r}(t),
\label{eq:reservoir_readout}
\end{equation}
where only the readout weight matrix $W_{out}$ is trained, via ridge regression with regularization coefficient $\beta_{rc}$:
\begin{equation}
W_{out} = \mathbf{Y}\mathbf{R}^\top\left(\mathbf{R}\mathbf{R}^\top + \beta_{rc}\mathbf{I}\right)^{-1},
\label{eq:ridge_regression}
\end{equation}
with $\mathbf{Y}$ and $\mathbf{R}$ denoting, respectively, the matrices of target outputs and corresponding reservoir states collected over the training trajectory. Figure~\ref{fig:fig2} summarizes this input$\to$reservoir$\to$output structure. The internal reservoir connections are kept fixed throughout training, and only the output weights are optimized.

This architecture has previously been applied with good results to chaotic time-series prediction \cite{ref14} and, more recently, to model-free forecasting of large, high-dimensional spatiotemporally chaotic systems \cite{ref15}. What is less established is how well it performs on a real mechanical system with a non-smooth, contact-induced nonlinearity, such as the jumping quarter-car model considered here, as opposed to a smooth textbook chaotic system such as the Lorenz or R\"ossler systems. This is the question addressed in this report.

\section{Results and Discussion}

Reservoir computing is applied here to predict three aspects of the model: bifurcation structure, attractor topology, and time-series behavior. The vehicle parameters are set to $\alpha = 2.333$, $\beta_j = 10$, $\zeta = 0.1552$, $\gamma = 0.113$, and $\Omega = 6.8$. The echo state network parameters are set to $N_{rc} = 1350$, $\alpha_{rc} = 0.75$, $\rho_{rc} = 0.547$, $k_{rc} = 24$, $W_{in} = 0.185$, and $\beta_{rc} = 7\times10^{-6}$. These values were determined by Bayesian optimization of the validation prediction error, followed by manual fine-tuning.

\subsection{Prediction of bifurcation}

\begin{figure}[htbp]
\centering
\includegraphics[width=0.55\textwidth]{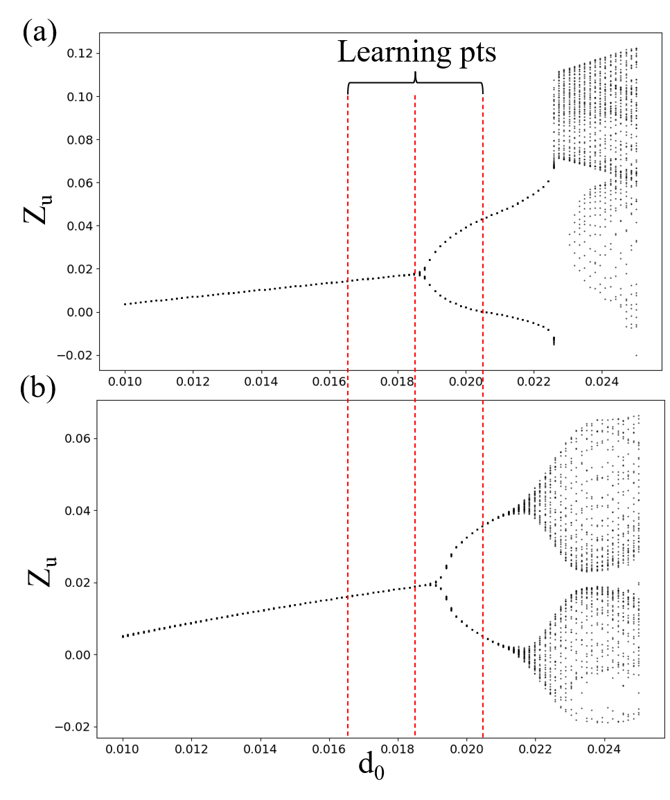}
\caption{Bifurcation diagrams of the tire deflection $Z_u$ as a function of the road excitation amplitude $d_0$, comparing (a) the ground-truth response obtained by direct numerical integration of the governing equations with (b) the response reconstructed by the trained echo state network. Vertical dashed lines in panel (a) mark the three operating points ($d_0 = 0.0165$, $0.0185$, and $0.0205$) used to train the reservoir. The predicted diagram in panel (b) qualitatively reproduces the period-doubling route to chaos observed in the ground truth, including extrapolation beyond the training range into the chaotic regime.}
\label{fig:fig3}
\end{figure}

Bifurcation is one of the major instabilities in nonlinear dynamics. Figure~\ref{fig:fig3}(a) shows the ground-truth bifurcation diagram of $Z_u$ for the quarter-car model with jumping nonlinearity with respect to $d_0$. Period-1 motion ($P_1$) is observed from $d_0 = 0.01$ to $0.0188$, after which period-2 motion ($P_2$) appears. A discontinuous bifurcation occurs and chaotic motion (C) emerges at $d_0 = 0.0225$. To predict the bifurcation, the echo state network is trained at three operating points: $d_0 = 0.0165$ ($P_1$), $d_0 = 0.0185$ ($P_1$), and $d_0 = 0.0205$ ($P_2$). Figure~\ref{fig:fig3}(b) shows the predicted bifurcation diagram. The predicted diagram is qualitatively similar to the ground truth: period doubling occurs at $d_0 = 0.0194$, followed by period-2 motion that becomes chaotic at $d_0 = 0.022$. This transition appears continuous in the prediction, whereas the ground-truth bifurcation is discontinuous -- likely because the reservoir's smooth, continuous-state dynamics tend to average across the abrupt switching behavior near the contact-loss discontinuity, rather than reproducing it exactly. Despite this discrepancy, the reservoir predicts the overall bifurcation structure qualitatively well, including extrapolation beyond the training range into the chaotic regime.

\subsection{Prediction of attractors}

\begin{figure}[htbp]
\centering
\includegraphics[width=\textwidth]{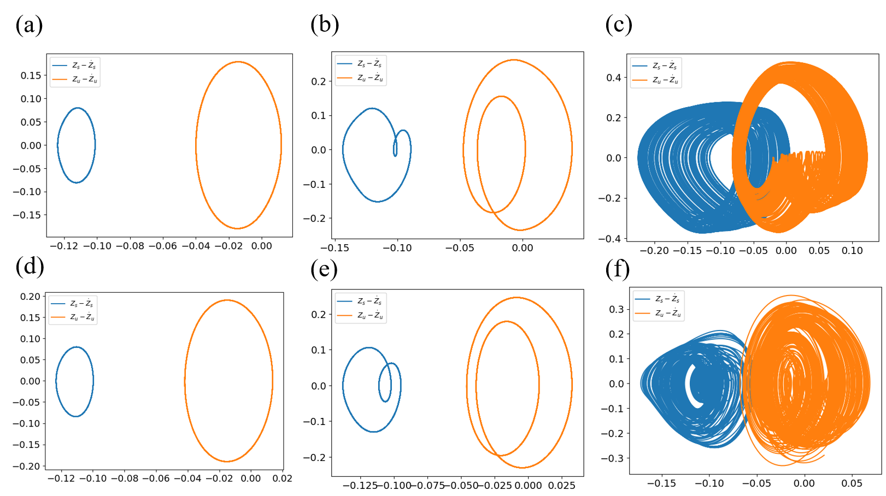}
\caption{Comparison of ground-truth and reservoir-predicted phase-space attractors, projected onto the $(Z_s, \dot{Z}_s)$ and $(Z_u, \dot{Z}_u)$ planes, for three representative dynamical regimes: period-1 motion at $d_0 = 0.015$ (a, d), period-2 motion at $d_0 = 0.02$ (b, e), and chaotic motion at $d_0 = 0.025$ (c, f). Panels (a)--(c) show the ground-truth attractors obtained from direct simulation; panels (d)--(f) show the corresponding attractors reconstructed from the trained reservoir's autonomous output. Close qualitative agreement in loop structure and, for the chaotic case, folded-sheet topology, indicates that the reservoir captures the underlying geometric organization of the flow rather than merely its instantaneous amplitude.}
\label{fig:fig4}
\end{figure}

To evaluate the topological accuracy of the prediction, Figure~\ref{fig:fig4} compares ground-truth and predicted attractors. Panels (a)--(c) and (d)--(f) show the ground-truth and predicted attractors, respectively, for $P_1$ ($d_0 = 0.015$), $P_2$ ($d_0 = 0.02$), and C ($d_0 = 0.025$). The predicted attractors closely reproduce the qualitative loop structure of the periodic cases and the folded-sheet topology of the chaotic case. For all training runs, ground-truth trajectories were generated by numerical integration of Eqs.~(\ref{eq:Zs_eom})--(\ref{eq:Zu_eom}) using a fourth-order Runge--Kutta scheme with a fixed time step of $0.01$. Training trajectories consisted of $500$ time units at each operating point, following a transient of $10^5$ time units that was discarded to remove initialization effects. 

\subsection{Prediction of time series}

\begin{figure}[htbp]
\centering
\includegraphics[width=\textwidth]{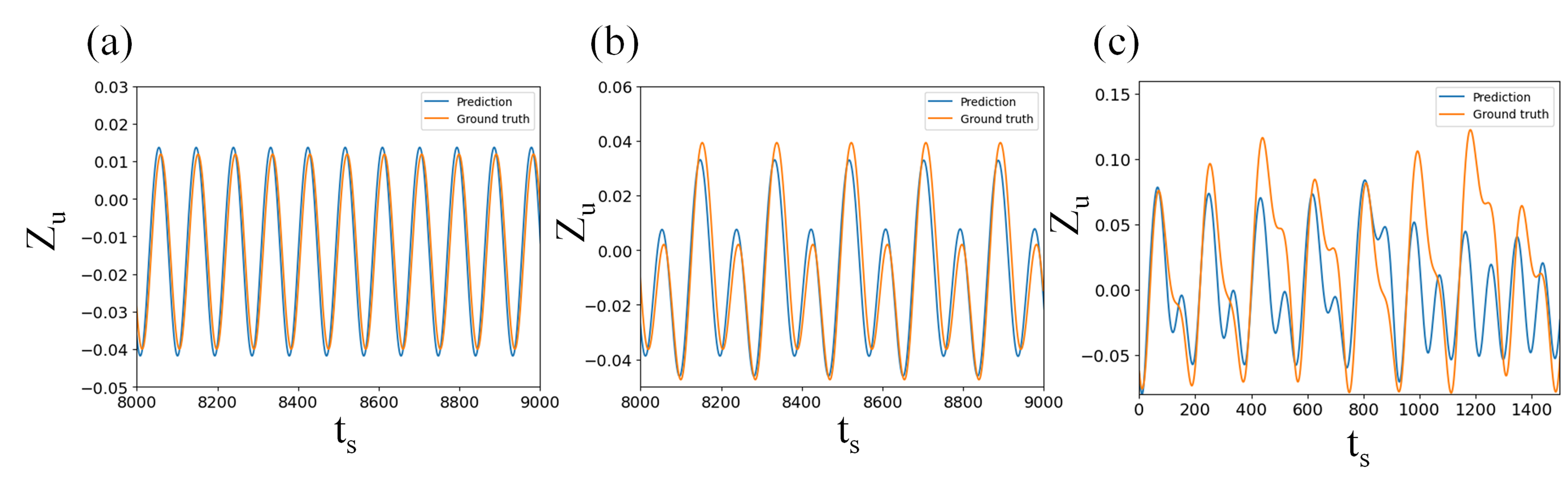}
\caption{Time-domain forecasts of the suspension stroke $Z_s$ generated by the trained echo state network (blue) superimposed on the corresponding ground-truth trajectories (orange), for (a) period-1 motion at $d_0 = 0.015$, (b) period-2 motion at $d_0 = 0.02$, and (c) chaotic motion at $d_0 = 0.025$. The reservoir sustains accurate long-horizon tracking of the periodic waveforms in (a) and (b), whereas in the chaotic case (c) the predicted and ground-truth trajectories diverge after a comparatively short horizon, consistent with the exponential error growth expected from sensitive dependence on initial conditions.}
\label{fig:fig5}
\end{figure}

Time-series prediction is the most important outcome for practical use, since it can be directly applied to predictive control of the vehicle. Predictions were generated using the trained reservoir for each type of motion. Figure~\ref{fig:fig5}(a)--(c) shows predicted time series of $Z_s$ for $P_1$ ($d_0 = 0.015$), $P_2$ ($d_0 = 0.02$), and C ($d_0 = 0.025$), respectively. Long-term prediction is achieved for $P_1$ and $P_2$, indicating that the reservoir accurately captures periodic dynamics. For the chaotic case, only short-term prediction is possible due to sensitive dependence on initial conditions -- a defining characteristic of chaotic motion -- after which the predicted and ground-truth trajectories diverge exponentially, consistent with expected behavior near a positive leading Lyapunov exponent.

\section{Conclusion}

In this work, the feasibility of reservoir computing for prediction of a nonlinear vehicle model was investigated through numerical simulations. Trained on data from only three operating points, the reservoir qualitatively predicted the bifurcation structure, attractor topology, and time-series behavior of a jumping quarter-car model across period-1, period-2, and chaotic regimes, including reasonable extrapolation beyond the training range. These results support reservoir computing as a viable data-driven predictor for practical mechanical systems with non-smooth, contact-induced nonlinearities. Future work should include quantitative validation via Lyapunov analysis and valid prediction time, as well as extension to predictive control applications.

\bibliographystyle{unsrt}
\bibliography{references}

\vspace{1em}
\noindent \textbf{Funding sources}

\noindent This work was supported by Japan Society for the Promotion of Science Grant-in-Aid (Nos. 26K18165).

\vspace{0.5em}
\noindent \textbf{Declaration of Competing Interest}

\noindent The authors have no relevant financial or non-financial interests to disclose.

\vspace{0.5em}
\noindent \textbf{Data Availability}

\noindent The datasets generated during the current study are available from the corresponding author on reasonable request.

\end{document}